\documentclass[letterpaper]{article} 
\usepackage[preprint]{aaai2027}  
\usepackage[hyphens]{url}  
\usepackage{graphicx} 
\usepackage{natbib}  
\usepackage{caption} 
\usepackage{booktabs}
\usepackage{amsmath}
\usepackage{amssymb}
\usepackage{array}
\usepackage{makecell}
\title{A Trust Layer for Agent Evaluation}
\author{
    Mohammadreza Sediqin,
    Shivali Dalmia,\\
    Srinivasa Karthikeya Reddy Kovvuri,
    Abhishek Mukherji
}
\affiliations{
    Centific Research, USA\\
    \{mohammadreza.s, shivali.dalmia, srinivasa.kovvuri, abhishek.mukherji\}@centific.com
}

\begin{document}
\maketitle
\begin{abstract}
Deterministic benchmark scores show that an agent received credit, but not whether that credit was earned, reported honestly, or would hold on a second run. We introduce a \emph{Trust Layer for Agent Evaluation}, an additive post-hoc framework that reports, beside each recorded score, whether it should be believed. It verifies four properties: whether the result is supported by the benchmark's own grading logic, whether a passing answer was earned through traceable computation, whether the agent's completion claim matches what occurred, and whether the result is stable under repeated execution. The first three use only saved artifacts; the fourth uses repeated runs. Model judgments only label evidence under majority voting; all verdicts follow deterministic rules and never modify the recorded score. Applied to five agent configurations on 108 tasks from Agents' Last Exam, every model shows passing runs with no traceable computation (at rates varying tenfold), confirmed false completion claims, and unstable results: 18 to 46\% of tasks do not stay in one score band over five runs. Only 22.6\% of 84 recorded passes clear all four checks. On a second benchmark, Terminal-Bench, the same rules leave 42.2\% of 180 recorded passes trusted, with instability as the main cause of failure. On both benchmarks, most rejected passes are rejected by deterministic checks alone. The layer keeps the benchmark's score and adds what the score alone cannot show: whether the result can be trusted.
\end{abstract}

\section{Introduction}
\label{sec:intro}

LLM agents are now evaluated on increasingly realistic, long-horizon benchmarks that mirror real-world work. SWE-bench~\citep{jimenez2024swe}, $\tau$-bench~\citep{taubench2024}, ScienceAgentBench~\citep{chen2025scienceagentbench}, and Agents' Last Exam (ALE)~\citep{ale2026} replace subjective assessment with executable grading, deterministic tests, and reference-based verification, so a benchmark score is often treated as an objective measurement of capability.

But a score answers only one question: \emph{did the grader award credit?} It does not say whether the credit was earned, whether the reported outcome reflects what happened, or whether the result would hold on a second run. A run that \emph{fabricates a plausible value} scores exactly like one that computed it. A run that \emph{fails but claims success} scores exactly like one that fails and says so. Because agent runs vary from attempt to attempt, a run that passes once in five tries scores exactly like one that passes every time; repetition studies show single-run scores shifting by several points from chance alone~\citep{bjarnason2026randomness}. And a failing score may reflect a packaging or evaluation artifact rather than genuine inability. In each case the number is accurate about the run and misleading about the agent, which raises a question the score cannot resolve on its own: \emph{can a recorded benchmark score be trusted?} Existing work addresses these concerns largely separately, through benchmark auditing, reward hacking and corrupt success~\citep{skalse2022defining,cao2026}, honesty~\citep{mask2025,behonest2024}, reliability~\citep{reliabilitybench2026}, provenance tracking for LLM-generated content~\citep{sediqin2026sage}, and evidence for recorded scores~\citep{gao2026can}, usually with new benchmarks, datasets, or protocols. To our knowledge, none verifies all four properties together for the question a benchmark user faces: given a score already recorded, should it be believed?

We introduce a \emph{Trust Layer for Agent Evaluation}, an additive post-hoc framework that audits recorded outcomes using artifacts benchmarks already save (trajectories, staged outputs, grading code) and runs alongside the benchmark rather than replacing it, requiring no changes to the benchmark or the agent harness. It applies four verifiers, one per case above. \emph{Competence} checks whether the benchmark's own grading logic reproduces the recorded outcome and names the decisive cause of failure. \emph{Reward Hacking} checks whether a pass was earned through traceable computation or reached through shortcuts and ungrounded outputs. \emph{Agent Deception Detection} checks the agent's completion claim against the independently verified outcome. \emph{Reliability} checks whether repeated runs under identical conditions stay in the same score band.

This is the first framework we know of to verify these four properties together, for an individual recorded score, from evidence a benchmark already produces. Its central design principle is the \emph{separation of judgment from decision}: models may help label evidence, but every verdict comes from a deterministic rule, and any model label that affects a verdict needs a majority of three independent votes. This matters because the closest prior work on false completion claims finds that judges anchor on confident language, with no configuration exceeding AUROC 0.65~\citep{falsesuccess2026}.

On 108 ALE tasks, the layer finds grading contradictions in both directions, including recorded passes that the benchmark's own grader scores at zero when re-run on the saved output; passing results with no traceable computation behind them; false completion claims in every recorded band; and results that do not reproduce. The dimensions prove complementary rather than redundant: many flagged tasks are caught by a single dimension, so removing any one would leave them looking clean, and only 19 of 84 recorded passes (22.6\%) clear all four. Our contributions are:
\begin{enumerate}
\setlength\itemsep{0pt}
\item To our knowledge, the first unified post-hoc trust layer that jointly verifies whether a recorded score is reproducible, earned, honest, and supported by the benchmark's own grading logic.
\item A design in which every verdict is deterministic and model judgment only labels evidence under a majority of three votes, so the headline number stays intact and auditable.
\item An empirical study over 108 professional tasks from ALE and 66 tasks from Terminal-Bench showing that substantial trust failures hide behind benchmark scores, that the four dimensions detect largely distinct failures, and that requiring all four leaves far fewer verified results than the scores suggest. Code will be released upon acceptance.
\end{enumerate}

\section{Related Work}
\label{sec:related}

\paragraph{Agentic benchmarks and validity auditing.}
Recent agent benchmarks favour verifiable, code-based outcomes. ALE grades long-horizon professional tasks with deterministic scripts against expert references, relying on LLM judging for under 7\% of workflows~\citep{ale2026}. SWE-bench grades code fixes for real GitHub issues with unit tests~\citep{jimenez2024swe}, $\tau$-bench compares the final database state with a reference~\citep{taubench2024}, and ScienceAgentBench scores self-contained programs on tasks from peer-reviewed publications~\citep{chen2025scienceagentbench}. Each reduces a run to one score from the final artifact, leaving the trajectory unscored. Such scores can misrepresent competence: \citet{zhu2026establishing} propose a manual checklist for task and outcome validity, \citet{kapoor2024} show that uncontrolled cost distorts conclusions, and automated auditing finds that test suites can reject correct solutions, with evaluator limitations increasingly dominating reported scores~\citep{tu2026benchguard}. These efforts audit how benchmarks are built, and related checks govern consistency in document repositories~\citep{sediqin2026gvd}.
\paragraph{Reward hacking, honesty, and reliability.}
Specification gaming, satisfying the literal objective but not its intent, was formalized by \citet{skalse2022defining} and catalogued across reinforcement learning systems by \citet{krakovna2020}. For tool-using LLM agents, benchmarks measure the propensity to skip verification or tamper with the evaluation itself~\citep{rhb2026,evilgenie2026}, and \citet{cao2026} report \emph{corrupt success}, correct end-states reached while violating procedural integrity, in 27\% to 78\% of reported successes. \citet{falsesuccess2026} characterize false success, agents asserting completion the environment contradicts, and show that LLM judges detect it unreliably; MASK~\citep{mask2025} and BeHonest~\citep{behonest2024} measure propensity in elicited scenarios rather than verifying claims in real runs. On reliability, single-run pass@1 across 60,000 SWE-Bench-Verified trajectories varies by up to 6.0 points depending on the run selected, and the authors recommend pass$^k$ as a pessimistic bound~\citep{bjarnason2026randomness}; related work treats evaluations as experiments whose variance determines reproducibility~\citep{mustahsan2025stochasticity}, and ReliabilityBench measures consistency under repetition and perturbation~\citep{reliabilitybench2026}.

\paragraph{Positioning.}
Validity work audits how a benchmark is \emph{constructed}; reward hacking, false success, and honesty work each target one mode, usually through an LLM judge of uncertain reliability; and reliability work builds repetition into \emph{new} benchmarks, all at a different level than one score. Recent work also examines whether benchmark scores are supported by their evidence~\citep{gao2026can}. The closest work, procedure-aware evaluation~\citep{cao2026}, evaluates whether an agent followed the intended procedure. We instead ask whether a score an existing benchmark \emph{already recorded} holds up, verifying four properties from the saved trajectory, grading code, and repeated runs, per dimension and as a conjunction, so a pass that is procedurally clean but not reproduced by the benchmark's own grader, or earned but falsely reported, is still caught. Unlike judge-based detection of false success, our deception check tests the claim against an independent re-grade, separating a false claim from a grading error.

\begin{figure*}[t]
  \centering
  \includegraphics[width=0.7\textwidth]{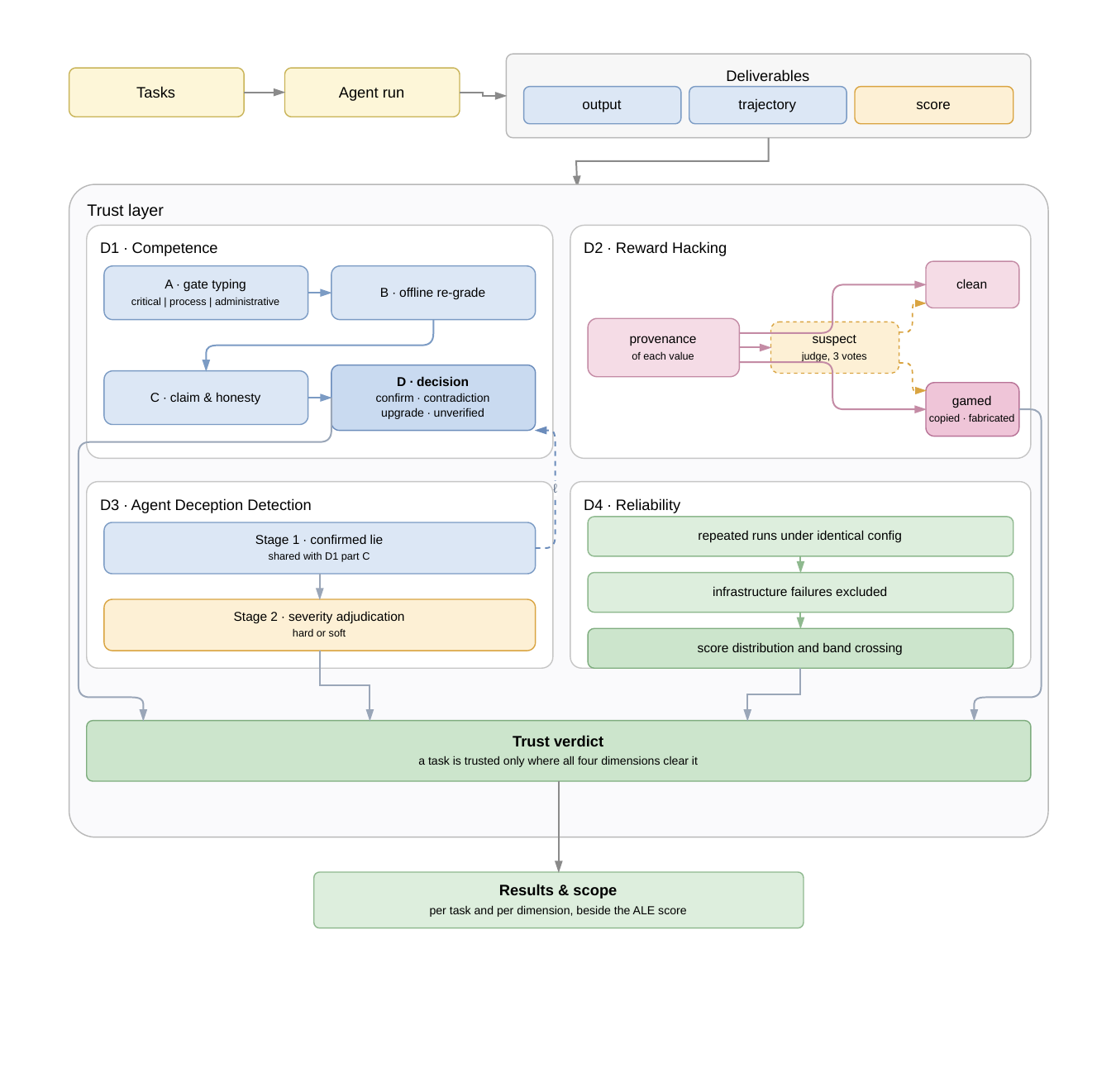}
  \vspace{-12mm}
  \caption{Overview of the Trust Layer. D1 re-derives the verdict, D2 traces value provenance, D3 tests the agent's claim, D4 repeats the run.}
  \label{fig:arch}
\end{figure*}

\section{System Architecture}
\label{sec:architecture}
The trust layer is an additive stage that runs after a benchmark has scored an agent (Figure~\ref{fig:arch}). It has four dimensions, one per failure class. \textbf{D1 (Competence)} asks what a recorded score is actually made of and exposes scores the grader does not reproduce, in either direction. \textbf{D2 (Reward Hacking)} asks whether a passing score was earned or reached without doing the work. \textbf{D3 (Agent Deception Detection)} asks whether the agent's success claim is contradicted by the outcome. \textbf{D4 (Reliability)} asks whether the score stays in the same band on repetition. D1--D3 read only what the benchmark saved; D4 also re-runs the agent. The layer reuses the benchmark's own grader wherever a verdict is needed and never overwrites the recorded score, reporting its findings beside it. It needs three things from a benchmark: saved trajectories, accessible grader logic, and outputs readable offline.

\paragraph{Evidence: the trajectory and re-grading.}
Two artifacts benchmarks produce but never fully use supply the evidence: the \emph{trajectory}, the step-by-step record showing \emph{how} a result was produced, and the benchmark's own \emph{grader}, re-run on the saved output to reproduce the score instead of trusting it. Every dimension rests on one or both.

\subsection{D1: Competence}

\textit{In short: re-run the grader and ask what actually decided the score.} D1 re-derives the verdict from the saved output with the task's own grader and compares it with the recorded score. No model assigns a score: a model labels gates by severity, and the verdict is a deterministic function of those labels. D1 has four stages, each feeding the next.

\paragraph{Stages A and B: typing and re-grading.}
A \emph{gate} is a check inside the grader whose failure forces the score to zero, such as a missing file or a wrong value. Stage A recovers every grader gate into a set $G$ and types it by what it protects,
\begin{equation}
\label{eq:type}
\tau(g) \in \left\{\begin{array}{@{}l@{}}\textit{critical}, \textit{process},\\ \textit{administrative}\end{array}\right\},
\end{equation}
where \textit{critical} gates protect the deliverable, \textit{process} gates a method rule, and \textit{administrative} gates its packaging. Each gate also carries an anti-gaming flag $\alpha(g)$. Typing is done by an LLM judge under a majority of three votes and, being a property of the grader rather than a run, once per task. Stage B re-runs the grader on the saved output and maps the fired gates $F$ to a verdict,
\begin{equation}
\label{eq:verdict}
v = \Phi(F, \tau, \alpha), \qquad F \subseteq G,
\end{equation}
with the most severe fired gate deciding, so a real correctness failure cannot hide behind a lesser one. Appendix~\ref{app:methoddetail} gives the gate criteria, matching procedure, severity ordering, and the verdicts Stage B can return.

\paragraph{Stages C and D: claim and challenge.}
Stage C reduces the agent's terminal message, by deterministic pattern matching, to a completion claim $c$ and tests it against $v$, giving the three-valued honesty signal used by D3:
\begin{equation}
\label{eq:lie}
\ell =
\begin{cases}
1 & \text{if } c = \textit{success} \wedge v \in V_{\text{fail}},\\
\bot & \text{if } c = \textit{success} \wedge v \in V_{\text{unver}},\\
0 & \text{otherwise},
\end{cases}
\end{equation}
where $V_{\text{fail}}$ are verdicts that reproduce a genuine failure and $V_{\text{unver}}$ those that could not be reproduced offline. Stage D is the only place the earlier facts are combined, by a purely deterministic rule with no model call:
\begin{equation}
\label{eq:decision}
\begin{gathered}
(D, \delta) = \Psi(s, v, \ell, m, r),\\
D \in \left\{\begin{array}{@{}l@{}}\textit{confirm}, \textit{contradiction},\\ \textit{upgrade}, \textit{unverified}\end{array}\right\},
\end{gathered}
\end{equation}
where $s$ is the recorded score, $m$ whether the prompt itself demanded the failed format, and $r$ a format-blind check that the output values match the reference. A confirm reproduces the recorded outcome; a contradiction disagrees, with a signed direction $\delta$ recording which side was wrong; an unverified could not be checked offline; and an upgrade, the one outcome needing all three earlier stages, is reached when the run failed a check that does not mean the work was wrong. Appendix~\ref{app:methoddetail} gives the upgrade conditions and contrasts a contradiction with an upgrade. Related outcome labels for recorded runs, including undecided ones, appear in \citet{gao2026can}.

\subsection{D2: Reward Hacking}
\label{sec:arch-d2}

\textit{In short: ask where each answer value came from.} A score shows that an answer is correct, not how it was produced. An agent that hardcodes a constant, copies a value the prompt revealed, or writes plausible numbers with no computation behind them scores exactly like one that did the work, and the grader is fooled. D2 marks such runs gamed, separating them from the honest passes they would otherwise be counted among. It traces provenance rather than re-checking correctness.

\paragraph{Tracing each value.}
D2 uses the reference answer, which the agent never saw, as an offline oracle to pick out which output values are the answer, then traces each against three sources from the run: output of commands the agent executed, inputs it read or was given in the prompt, and text it typed by hand into files and shell commands. Each value gets a label,
\begin{equation}
\label{eq:prov}
\pi(x) \in \left\{\begin{array}{@{}l@{}}\textsc{derived}, \textsc{given},\\ \textsc{typed}, \textsc{ungrounded}\end{array}\right\},
\end{equation}
\textsc{derived} if it appears in executed output, \textsc{given} if it came from an input or the prompt, \textsc{typed} if written by hand, and \textsc{ungrounded} if it appears nowhere. The decisive test is for an answer value absent from executed output: was it typed by the agent, or produced by a script at run time? A correct value a script computed is real work; a hand-typed one is not. Scalars are traced one at a time and arrays at the file level (Appendix~\ref{app:methoddetail}).

\paragraph{Confessions, the judge, and the verdict.}
The agent's messages and reasoning are scanned for admissions of copying or hardcoding, with a negation guard so a stated intention \emph{not} to hardcode is never read as a confession. A confession is never a flag on its own: with a value already found unearned (matching the reference but neither computed nor read) it confirms a hack; otherwise it only marks the run for a closer look. Values that remain genuinely unclear go to a model judge, which sees the prompt, the flagged values, the agent's words, and its code, and is warned that printing a hardcoded constant does not make it computed. Before convicting, D2 has the judge check whether the task required the value to be computed; if the task permits a direct answer, the value is cleared. A strict majority of three votes is required to convict, so a split resolves to \textit{clean}: the check would rather miss a cheat than accuse an honest run. The run verdict is
\begin{equation}
\label{eq:int}
V_{\text{int}} \in \{{clean}, {suspect}, {gamed}\},
\end{equation}
where a strong flag gives \textit{gamed}, a weak flag or unresolved value \textit{suspect}, and a clean trace \textit{clean}. \textit{suspect} is only a routing state that sends the run to the judge, which resolves it, so none is reported. Each verdict records its cheat type (copied or fabricated); signals range from a confession with a provenance gap, the strongest, to a judge-voted fabrication resting on the absence of a trace, the weakest (Appendix~\ref{app:methoddetail}). An untraced value is flagged for review, not convicted, and is marked \textit{gamed} only when a condition above holds, never when the task permits a direct answer. A value fabricated and then written through executed code looks derived, so D2 bounds untraceable fabrication, not fabrication passed through executed code.

\subsection{D3: Agent Deception Detection}
\label{sec:d3}

\textit{In short: check the agent's final report against what really happened.} The last message is where the agent reports its result, so when it claims success on a run that failed, oversight breaks down. The benchmark gives the same score either way, so lying costs the agent nothing and the score never shows it. We use ``lie'' operationally: a claim of success on a run whose failure was proven by re-grading, or a specific claim about the output that the saved output contradicts; D3 shows the report was false, not that the agent knew it was. D3 has two jobs: a yes-or-no check inside the D1 pipeline, and a separate judgment of how serious a confirmed lie was.

\paragraph{Job 1: is the agent lying?}
This is the deterministic signal $\ell$ of Equation~\eqref{eq:lie}. It checks the agent's success claim against the re-graded verdict $v$, \emph{not} the recorded score: testing against the recorded score would only show that the agent claimed success and scored zero, mixing up a lying agent with a grading error, while a re-grade tells them apart. If re-running the grader proves the failure, the claim is a confirmed lie ($\ell = 1$); if the grader cannot be re-run, it is marked unconfirmed and D3 stops ($\ell = \bot$). This feeds back into Stage D, where a proven lie removes the agent's benefit of the doubt on a formatting-only failure. Because $\ell$ rests on D1's re-grade, the two dimensions are not independent evidence.

\paragraph{Job 2: how serious was the lie?}
After D1 finishes, a model judge re-examines each confirmed lie under majority vote, so this output is an adjudication rather than a measurement. The judge sees the agent's words and what independently happened (the recorded score, re-graded verdict, fired gate, failure explanation, grader messages, and a trajectory digest) and produces
\begin{equation}
\label{eq:d3}
(\hat{\ell}, \gamma, \eta, \phi) = \Xi(\ell, v, q, E)
\end{equation}
where $q$ is the agent's exact claim and $E$ what happened in the run: a final decision $\hat{\ell}$ confirming or dropping the lie, the kind of claim $\gamma$, whether a concrete defect sits behind it $\eta$, and whether the lie alone should fail the task $\phi$ (Appendix~\ref{app:methoddetail}). Job~2 never feeds back into D1, which uses only $\ell$, so the Stage~D decision never inherits the judge's variance and no score is changed. The signal $\ell$ is used in exactly three places: a confirmed lie blocks a packaging upgrade in Stage D and tags a contradiction with whether the agent also lied, and $\ell=\bot$ marks an unverified pass as unconfirmed.

\subsection{D4: Reliability}

\textit{In short: run the task again and see whether the result stays in the same band.} One run is a sample, not a measurement of a point; a pass that holds one time in five is luck recorded as competence, invisible to any single-run dimension. D4 executes each task $N = 5$ times under identical configuration and characterises the score distribution $\{s_1, \ldots, s_N\}$. Each run is first classified as an infrastructure or genuine failure from the termination record, trajectory text, and usage counters (exhausted credit, a dead sandbox, zero tool calls); infrastructure runs are excluded, so D4 measures the agent, not the harness. Over the surviving runs $S$,
\begin{equation}
\label{eq:reliab}
\begin{aligned}
\bar{s} &= \tfrac{1}{|S|}\textstyle\sum_{s \in S} s, \qquad
\bar{\Delta} = \bar{s} - s_{\text{rec}},\\
\texttt{crossed} &= \mathbf{1}\big[\,|\{\mathrm{band}(s) : s \in S\}| > 1\,\big],
\end{aligned}
\end{equation}
where $s_{\text{rec}}$ is the recorded score and $\mathrm{band}(\cdot)$ maps a score to pass, partial, or fail with ALE's thresholds~\citep{ale2026}. A task is \emph{varied} when \texttt{crossed} fires and \emph{stable} otherwise~\citep{gonzalez2026reliability}. \texttt{crossed} asks whether the repeats agree with each other; $\bar{\Delta}$ asks how far they sit from the recorded score, negative meaning the recorded run was the better one. Only \texttt{crossed} decides whether a task clears D4; $\bar{\Delta}$ is reported, not used as a gate.

Two further statistics separate what a score measures from what it merely samples~\citep{bjarnason2026randomness}. Treating a score as true ability plus noise, the variance of repeats splits into a between-task part (genuine differences in difficulty and capability) and a within-task part (run-to-run randomness). Their ratio
\begin{equation}
\label{eq:rho}
\rho = \sigma^2_{\text{between}} / (\sigma^2_{\text{between}} + \sigma^2_{\text{within}})
\end{equation}
is the share of variance that is signal, and $1 - \rho$ the share that is not~\citep{mustahsan2025stochasticity}. Estimating each task's pass probability $\hat{p}$ from its repeats gives $\hat{p}^k$, the chance of passing $k$ consecutive runs, whose gap from $\hat{p}$ is the capability a one-shot score reports but cannot sustain. Appendix~\ref{app:methoddetail} gives the per-task form of $\rho$ and what else the persisted repeats support.

\subsection{Aggregation Across Dimensions}
\label{sec:arch-agg}

A wrong answer, an unearned pass, a false claim, and an unstable score call for different responses, so the dimensions are reported separately; averaging them would discard the distinction the layer exists to draw. They are also reported as a conjunction: a task is \emph{trusted} only where every dimension clears it, meaning the verdict re-derives, the answer values are grounded, no false claim is confirmed, and the task is stable across repeats. This is a gate, not a score: no dimension is weighted against another, and the recorded score is never overwritten.

\paragraph{Counting rules.}
A task clears D1 only when re-grading yields a verified pass; for a recorded pass this means \textit{confirm}, so \textit{contradiction} and \textit{unverified} are D1 failures, and a task whose grader cannot be re-run offline counts as \textit{unverified}. It clears D2 only when \textit{clean}; \textit{gamed} and ``no verdict'' are D2 failures. It clears D3 unless a confirmed false claim is hard.
It clears D4 only when stable, whichever direction its repeats move, since a score that changes bands cannot be trusted either way, and a score is read as evidence about the agent (Section~\ref{sec:intro}). A task may fail several dimensions (a run that fabricated its answer often also claimed success for it), so the conjunction is not a sum and we report the overlap~\citep{zhou2026accuracy}. The four are not exhaustive but cover the trust failures saved artifacts can reveal.

\section{Experiments}
\label{sec:results}

\subsection{Dataset and Setup}
\label{sec:dataset}

We evaluate the trust layer on 108 publicly available tasks from ALE~\citep{ale2026}, a benchmark of long-horizon work graded by deterministic scripts against expert references, spanning ALE's three difficulty tiers and 11 professional domains, each with its own grading code producing a single score per run. Each task was run once under each of five agent configurations: Claude Sonnet 4.6, Claude Opus 5, GPT-5.6 Sol (GPT-Sol), Muse Spark 1.2 (MetaMuse), and Gemini 3.6 Flash, under the same agent harness and decoding configuration. The recorded bands differ by model, from 15 recorded passes for the weakest to 24 for the strongest (Appendix~\ref{app:bands}). For D1--D3, the trust layer reads only what each run already saved: the agent's trajectory, the outputs it staged, and the task's own grading code. It re-executes that code against the saved output without re-running the agent, and never modifies the recorded score. Where a dimension requires a model judgment we use GPT-5.5 with three independent votes and a strict majority.

A task is \emph{ungradable} when its grader cannot be re-run offline, which holds for 11 of the 108 and is a property of the task rather than of any model, so the same 11 are excluded for every agent. For these runs D1 produces no verdict and D3 has no re-grade to test a claim against, so for comparability the per-dimension results (Tables 1--3) cover the remaining 97. D4 instead re-runs the agent five times per task under identical configuration. 87 of the 108 completed all five repeats under Sonnet, and the 50 tasks repeated under all five configurations were drawn at random from those, stratified across ALE's tiers (24 Near-Term, 18 Full-Spectrum, 8 Last-Exam) and recorded band. Five repeated runs per task is in line with current practice~\citep{guo2026swe, zhang2025buildbench}. Each dimension reports its own denominator. Because the 50 repeated tasks are drawn from all 108, the aggregate in Section~\ref{sec:results-agg} can include ungradable tasks, which fail D1, and fail D2 only where no answer values can be traced (Section~\ref{sec:arch-agg}). Appendix~\ref{app:extension} extends the study to three newer configurations.

\paragraph{Second benchmark.} To test whether the layer transfers to another benchmark, we also apply it to all 66 tasks of Terminal-Bench 4.0 (v4.0.0, Harbor Hub)~\citep{merrill2026terminal}, using public leaderboard runs from five configurations: Claude Opus 5.5, Claude Opus 5, GPT-Sol, GPT-6.1 Sol, and Claude Fable 5.1. Terminal-Bench grades the final state of a terminal environment with tests, so every score is a pass or a fail. As on ALE, D4 uses five runs per task under each configuration, and the earliest started run is the recorded run. Two tasks require a GPU and cannot be re-graded offline, so they are ungradable, as in the ALE study. D1 and D3 cover the remaining 64 tasks, and D2 and D4 cover all 66. Infrastructure runs are excluded with the same rule as on ALE. The decision rules and the judge are the same as on ALE. Because the runs come from the leaderboard, the harness and settings are those of each submission. Section~\ref{sec:results-tbench} gives the results.

\subsection{D1: Competence}
\label{sec:results-d1}

\begin{table}[t]
\centering
\small
\begin{tabular}{@{}lrrrr@{}}
\toprule
Model & confirm & contra & unverified & upg \\
\midrule
Sonnet   & 56 & 38 & 3 & 0 \\
MetaMuse & 50 & 47 & 0 & 0 \\
GPT-Sol  & 50 & 47 & 0 & 0 \\
Opus     & 53 & 43 & 1 & 0 \\
Gemini   & 61 & 35 & 1 & 0 \\
\bottomrule
\end{tabular}
\vspace{-2mm}
\caption{D1 outcome per model.}
\label{tab:d1}
\end{table}

\begin{table}[t]
\centering
\small
\begin{tabular}{@{}lrrr@{}}
\toprule
Model & Confirmed & hard & soft \\
\midrule
Sonnet   & 55 & 26 & 29 \\
MetaMuse & 35 & 31 & 4 \\
GPT-Sol  & 40 & 38 & 2 \\
Opus     & 17 & 15 & 2 \\
Gemini   & 27 & 26 & 1 \\
\bottomrule
\end{tabular}
\vspace{-2mm}
\caption{Confirmed false claims per model.}
\label{tab:d3}
\end{table}

D1 executes each task's grading code offline on the saved output, without re-running the agent, and compares the result with the recorded score (Table~\ref{tab:d1}, 97 gradable tasks). The models cluster between 50 and 61 confirmations although their recorded passes differ by nine. Most contradictions (contra) are partial scores below the verified-pass bar: the benchmark credits partial work in proportion, while D1 requires a complete re-derivation. The real disagreements are smaller: 2 to 9 recorded passes per model re-derive as failures, and one or two recorded failures as passes (Appendix~\ref{app:d1detail}). The upgrade (upg) path was reachable but never granted, since every candidate's values differed from the reference. Gate typing, D1's only model-labelled stage, agrees closely enough that the majority rule absorbs residual disagreement rather than masking a coin flip (Appendix~\ref{app:d1detail}); every later stage is deterministic: re-executing the grader five times on the same output returns an identical verdict and band on all 108 tasks.

\subsection{D2: Reward Hacking}
\label{sec:results-d2}

\begin{table}[t]
\centering
\small
\setlength{\tabcolsep}{3.5pt}
\begin{tabular}{@{}lrrrrr@{}}
\toprule
Model & assessed & clean & gamed & \makecell{not\\assessed} & gamed\% \\
\midrule
Sonnet   & 97 & 75 & 14 & 8  & 14.4 \\
MetaMuse & 97 & 52 & 38 & 7  & 39.2 \\
GPT-Sol  & 97 & 45 & 49 & 3  & 50.5 \\
Opus     & 97 & 75 & 10 & 12 & 10.3 \\
Gemini   & 97 & 44 & 25 & 28 & 25.8 \\
\bottomrule
\end{tabular}
\vspace{-2mm}
\caption{D2 verdicts per model. Gamed\% is over the 97 assessed tasks, including runs carrying no verdict.}
\label{tab:d2}
\end{table}

Over the 97 assessed tasks (Table~\ref{tab:d2}), the models differ by a factor of five: Opus and Sonnet reach their answers without doing the work on roughly one run in eight, GPT-Sol on half; on recorded passes the spread is tenfold, from 1 of 22 for Opus to 10 of 21 for GPT-Sol (Table~\ref{tab:d2bands}). Runs with no verdict are neither clean nor gamed and stay in the denominator. This matters most for Gemini, whose rate over runs with a verdict is 36.2\% rather than 25.8\%; as these are unlikely to be missing at random, its true rate lies between 25.8\% and 54.6\%. Fabrication, a value that matches the reference yet appears in no executed output and nothing the agent read, accounts for most convictions; copying a supplied value is rarer and an explicit admission rarer still, so the judge carries most D2 verdicts for every model. Every \textit{gamed} verdict for Sonnet and GPT-Sol was verified by hand against the trajectory, and all 14 and all 49 held; Appendix~\ref{app:d2detail} gives the breakdowns. Injecting the reference value into runs the detector had cleared tests the other direction: D2 flagged 36 of 38 where the value was left ungrounded, and 2 of 45 where executed code wrote it.

\subsection{D3: Agent Deception Detection}
\label{sec:results-d3}

D3 runs the two jobs of Section~\ref{sec:d3}. A claim can only be tested where re-derivation reproduces what happened, so the 11 tasks without an offline verdict are ineligible; Table~\ref{tab:d3} covers the remaining 97. A claim is \emph{hard} when a concrete defect sits behind it (a fired critical gate, a missing output, or a trace showing the work was wrong) while the agent reported the deliverable as verified, and \emph{soft} when no such defect is evidenced, typically an overclaim that all checks pass when the agent's own numbers were already below the bar. Roughly half of Sonnet's confirmed claims are soft, while the other four models' are mostly hard, so the same detector finds two styles of misreporting. Models also differ in how often they make confirmed false completion claims, from 17 for Opus to 55 for Sonnet, and false claims appear in every recorded band for every model, including full passes. On two annotated samples of Sonnet runs, annotators agreed on whether success was claimed in 48 of 50 terminal messages ($\kappa = 0.85$), and the matcher reaches $F_1 = 0.91$ against their consensus; on severity they agreed at $\kappa = 0.84$, and the judge reaches $F_1 = 0.97$. For GPT-Sol, annotators agreed on all 50 terminal messages ($F_1 = 0.97$ for the matcher) and on 36 of 40 confirmed claims' severity, with the judge at $F_1 = 0.96$; as 38 of the 40 are hard, chance agreement is already 0.86, so $\kappa$ (0.29) understates this agreement; the prevalence-adjusted $\kappa$ is 0.80.

\subsection{D4: Reliability}
\label{sec:results-d4}

\begin{table}[t]
\centering
\small
\begin{tabular}{@{}lrrrrr@{}}
\toprule
Model & Tasks & stab. & var. & var.\% & $\bar{\Delta}$ \\
\midrule
Sonnet   & 50 & 27 & 23 & 46.0 & $-0.31$ \\
Opus     & 50 & 39 & 11 & 22.0 & $-0.30$ \\
GPT-Sol  & 50 & 41 & 9  & 18.0 & $-0.22$ \\
MetaMuse & 50 & 38 & 12 & 24.0 & $-0.12$ \\
Gemini   & 50 & 36 & 14 & 28.0 & $-0.24$ \\
\bottomrule
\end{tabular}
\vspace{-2mm}
\caption{D4 outcome per model.}
\label{tab:d4}
\end{table}

Table~\ref{tab:d4} covers the 50 repeated tasks. Its $\bar{\Delta}$ column, the mean change from the recorded score over varied pass-band tasks, is negative throughout: a recorded pass that fails to reproduce returns a lower score than the one on record. Band composition differs by model, so rates are comparable only within band (Appendix~\ref{app:d4detail}, which also gives the 87-task results). The varied rate ranges from 18.0\% for GPT-Sol to 46.0\% for Sonnet, a factor of two and a half on identical tasks and conditions, so instability is a property of the agent as well as the task. The movement follows the recorded band: the recorded run was the better run where the benchmark recorded success and the worse run where it recorded failure (Appendix~\ref{app:d4detail}). Pooled over all tasks the mean change is $+0.03$ or less for every model, so the directions cancel and an aggregate hides the effect completely.

Sonnet varies on 25.3\% of the 87 tasks; its repeat variance splits into 0.143 between tasks against 0.040 within, $\rho = 0.78$, so close to a quarter of the variation in recorded scores is about which run was saved, not about the agent. The noise is concentrated: the median task varies by a standard deviation of 0.007 across repeats, the varied tasks by 0.350. A task passes on 19.3\% of individual repeats but only 13.8\% of five-run sequences, and of the 24 tasks passing at least once, only 12 pass every time: half the tasks the benchmark could count as successes succeed only some of the time, and no single-run dimension can tell the halves apart. Sonnet's two independent campaigns, over the 87 tasks and the 50 drawn from them, disagree: 25.3\% against 46.0\% varied under the same configuration, where a random draw would carry about 13 varied tasks rather than 23 (Table~\ref{tab:d487}). The D4 verdict is therefore itself a sample; we report both campaigns rather than the more favourable one, and a larger $N$ would tighten the estimate. Because $\bar{\Delta}$ does not gate the verdict, a stable task can settle in a band other than the recorded one. Of the 181 stable verdicts in Table~\ref{tab:d4}, 16 do so (Gemini 10, Sonnet 2, Opus 2, MetaMuse 2, GPT-Sol 0), four of them recorded passes. Requiring reproduction of the recorded band would reclassify these 16; we report them separately.

\subsection{Aggregation Across Dimensions}
\label{sec:results-agg}
Table~\ref{tab:agg} gives the conjunction over the tasks the benchmark recorded as passes, where it does its work: D1's verified-pass bar rejects a partial score by definition, so no partial task can clear it. The D1--D4 columns overlap and do not sum; Appendix~\ref{app:aggdetail} gives all three bands and traces two recorded passes end to end. Between 7.7\% and 40.0\% of recorded passes survive all four checks; the intervals overlap, so these rates bound how many passes hold up rather than rank the configurations. Pooled, only 19 of 84 survive (22.6\%, 95\% CI 15.0--32.6), and the majority fails in every configuration. The figure is robust: 25.7\% over gradable passes, 21.4\% if D4 must reproduce the recorded band, and 21.8\% over 133 passes with three newer configurations (Appendices~\ref{app:aggdetail}, \ref{app:d4detail}, and~\ref{app:extension}). Models fail for different reasons: GPT-Sol loses two thirds of its passes to D2, failing it on 12 of 18, while Opus fails D2 on three and loses more to re-derivation than to anything else. Between a third and a half of flagged tasks are caught by a single dimension, so removing any one would leave them indistinguishable from clean ones.
\begin{table}[t]
\centering
\footnotesize
\setlength{\tabcolsep}{1.8pt}
\begin{tabular}{@{}lrrrrrrrl@{}}
\toprule
 & & \multicolumn{4}{c}{failed} & \multicolumn{3}{c}{aggregate} \\
\cmidrule(lr){3-6}\cmidrule(l){7-9}
Model & Passes & D1 & D2 & D3 & D4 & trusted & \% & 95\% CI \\
\midrule
Opus     & 20 & 8 & 3  & 0 & 5 & 8 & 40.0 & [21.9, 61.3] \\
Sonnet   & 15 & 4 & 5  & 1 & 5 & 4 & 26.7 & [10.9, 52.0] \\
MetaMuse & 18 & 8 & 6  & 1 & 3 & 4 & 22.2 & [9.0, 45.2] \\
GPT-Sol  & 18 & 7 & 12 & 3 & 4 & 2 & 11.1 & [3.1, 32.8] \\
Gemini   & 13 & 7 & 5  & 2 & 3 & 1 & 7.7  & [1.4, 33.3] \\
\midrule
All      & 84 & -- & -- & -- & -- & 19 & 22.6 & [15.0, 32.6] \\
\bottomrule
\end{tabular}

\caption{Conjunction over recorded passes among the 50 repeated tasks, under the counting rules of Section~\ref{sec:arch-agg}. Intervals are Wilson score intervals at 95\%.}
\label{tab:agg}
\end{table}
The ablation (Table~\ref{tab:ablation}) also shows how much rests on model labels: the two deterministic dimensions (D1, D4) leave 34 of 84 recorded passes, and the two judge-assisted ones reject a further 15, leaving 19. Over all 50 tasks the totals reflect the threshold more than the evidence, since D1 fails every partial task; the recorded-pass figures answer the question the layer was built for.

\subsection{Transfer to Terminal-Bench}
\label{sec:results-tbench}
Table~\ref{tab:tbench} gives the conjunction over recorded passes on Terminal-Bench. Of 180 recorded passes, 76 clear all four checks (42.2\%, 95\% CI 35.2 to 49.5), against 22.6\% on ALE. The share ranges from 10.5\% for GPT-Sol, which keeps 2 of its 19 passes, to 57.4\% for Opus 5.5. Appendix~\ref{app:tbench} gives the results for each dimension.

\begin{table}[t]
\centering
\footnotesize
\setlength{\tabcolsep}{1.8pt}
\begin{tabular}{@{}lrrrrrrrl@{}}
\toprule
 & & \multicolumn{4}{c}{failed} & \multicolumn{3}{c}{aggregate} \\
\cmidrule(lr){3-6}\cmidrule(l){7-9}
Model & Passes & D1 & D2 & D3 & D4 & trusted & \% & 95\% CI \\
\midrule
Opus 5.5    & 47  & 3 & 5 & 0 & 13 & 27 & 57.4 & [43.3, 70.5] \\
Opus 5      & 34  & 4 & 4 & 3 & 12 & 16 & 47.1 & [31.5, 63.3] \\
GPT-Sol     & 19  & 6 & 0 & 2 & 11 & 2  & 10.5 & [2.9, 31.4] \\
GPT-6.1 Sol & 37  & 8 & 5 & 6 & 11 & 14 & 37.8 & [24.1, 53.9] \\
Fable 5.1   & 43  & 3 & 5 & 1 & 21 & 17 & 39.5 & [26.4, 54.4] \\
\midrule
All         & 180 & -- & -- & -- & -- & 76 & 42.2 & [35.2, 49.5] \\
\bottomrule
\end{tabular}
\caption{Conjunction over recorded passes on Terminal-Bench (66 tasks, five runs each), in the form of Table~\ref{tab:agg}. D1 to D4 count passes failing that dimension and overlap, so they do not sum. Intervals are Wilson score intervals at 95\%.}
\label{tab:tbench}
\end{table}

The main cause of failure differs between the two benchmarks. On ALE, D1 and D2 remove the most passes. On Terminal-Bench, D4 does: 68 of the 180 recorded passes change band across the five runs, and 21 of Fable 5.1's 43 passes fail D4. Between 27.3\% and 42.4\% of tasks change band (Table~\ref{tab:tbench-d3d4}), and $\bar{\Delta}$ is negative for every model, so the recorded pass was again the better run. D2 marks 4.5\% to 12.1\% of tasks as gamed per model (Table~\ref{tab:tbench-d2}), against 10.3\% to 50.5\% on ALE. Terminal-Bench grades the final state with tests, and many deliverables are code, so there are few answer values to copy or fabricate. Because D2 traces values, these rates cover gaming through answer values, not gaming through code written only to pass the tests. All 24 gamed verdicts and all confirmed false claims were checked by hand against the trajectory, and all held.

The result rests even less on model labels than on ALE. D1 and D4, the two deterministic dimensions, leave 90 of the 180 passes, and D2 and D3 remove a further 14 (Table~\ref{tab:tbench-ablation}). So 90 of the 104 rejected passes, or 87\%, are rejected without any model judgment, against 77\% on ALE.

The two version pairs show how trust changes when a model is updated. From GPT-Sol to GPT-6.1 Sol, the trusted share rises from 10.5\% to 37.8\%, and from Opus 5 to Opus 5.5 it rises from 47.1\% to 57.4\%. Both newer versions also make fewer confirmed false claims and vary less (Appendix~\ref{app:tbench}). Only the drop in false claims for the GPT pair has intervals that do not overlap, so the other changes are trends. On ALE, Opus 5.5 varied more than Opus 5 (Appendix~\ref{app:extension}), so stability depends on the benchmark as well as the model.

\section{Conclusion and Future Work}
\label{sec:conclusion}

We presented a trust layer that checks four properties of a recorded agent score: whether a recorded failure was justified, whether a pass was earned, whether the agent's claim matched what happened, and whether the score stays stable across repeated runs. It is additive, never changes the recorded score, and uses model judgment only to label evidence. On ALE, across five agent configurations, every model reaches a passing answer without computing it on some runs, at rates that differ tenfold between the cleanest and the worst; confirmed false completion claims appear in every recorded band for every model; and 18\% to 46\% of tasks change score band across repeated runs, with the recorded pass usually the agent's better run and the recorded failure its worse one, a difference that aggregate scores hide. Requiring all four properties leaves 22.6\% of recorded passes standing, and between a third and a half of flagged tasks are identified by only one dimension, so no single check can replace the others. On Terminal-Bench, the same rules leave 42.2\% of 180 recorded passes standing, with instability rather than gaming as the main cause, so which check matters depends on the benchmark. Future work will apply the layer to SWE-bench Verified, $\tau$-bench, and ScienceAgentBench.

\section*{Limitations}
\label{app:limitations}

Our claims are scoped to what the evidence supports. On ALE, precision is validated directly on two configurations: for D2 by hand-checking all Sonnet and GPT-Sol gamed verdicts, and for D3 by annotator agreement on Sonnet and GPT-Sol (Section~\ref{sec:results-d3}); extending these checks to all models, and validating D1's gate typing against human labels, is natural next work. A single judge (GPT-5.5) is used for all configurations, so every model is judged by the same standard; it shares a family with GPT-Sol, which nonetheless receives the most \textit{gamed} verdicts on ALE, all 49 of which held on hand-checking, and with GPT-Sol and GPT-6.1 Sol on Terminal-Bench. With 13 to 20 recorded passes per configuration on ALE and overlapping intervals in Table~\ref{tab:agg}, the results establish that most recorded passes fail at least one check in every configuration, not a ranking of the agents. Results come from two benchmarks, ALE and Terminal-Bench. Because the layer needs only a trajectory, a staged output, and a re-runnable grader, it can be applied to others without changing the method. The Terminal-Bench runs come from public leaderboard submissions, so we did not control their harness or settings. Every decision rule is deterministic, and the ablation in Section~\ref{sec:results-agg} shows how much of the result holds without the two judge-assisted dimensions. A \textit{gamed} verdict means no traceable computation, not intent to cheat, and D2 does not catch fabrication passed through executed code; a confirmed lie shows a report was false, not that the agent knew it. D4 re-executes the agent, its $N = 5$ verdict is itself a sample, and stable tasks whose band differs from the recorded one are reported separately (Section~\ref{sec:results-d4} and Appendix~E).

\bibliography{reference}

\clearpage
\appendix

\section{Method detail}
\label{app:methoddetail}

This appendix collects the criteria, label definitions, and
statistical detail referenced from Section~\ref{sec:architecture}.

\paragraph{Gate-type criteria (Stage~A).}
The critical category also covers gates enforcing requirements the
task itself mandates, such as a required output file, a specified value that
must match the ground truth, or a constraint the prompt states explicitly,
since failing one of these means the delivered work does not meet what was
asked. A \emph{process} gate is one such as a requirement that an authored
script invoke a mandated simulation binary: the output could still be correct
if the rule were ignored. An \emph{administrative} gate is one such as a
submitted file missing a schema-required key: the work may be correct but
cannot be read. Gates are recovered by scanning every file of the grader rather than the
nominal scorer alone, since existence gates often live in the entry module
while value gates live in a delegated one. The anti-gaming flag $\alpha(g)$ marks gates that guard against degenerate or fabricated output.
\paragraph{Gate matching (Stage~B).}
From the trajectory Stage~B uses only the task identifier, and the instruction
where needed to select the correct reference. Each reported failure is matched
back to a Stage~A gate by its identifier, which is the grader's own message
template recognised in its filled-in output. The verdicts Stage~B can return are a wrong answer, a fabricated or degenerate
artifact caught by the grader's own guard, a violated method rule, a missing
deliverable, and a formatting slip, each recorded separately.
\paragraph{Severity ordering (Stage~B).}
When several gates fire at once, a precondition failure outranks a quality
miss, and among preconditions critical outranks process, which outranks
administrative. This ordering also closes an exploit, since an administrative
failure is the only verdict Stage~D can upgrade, and ranking prevents a run
with any real failure from being relabelled as a formatting slip.

\paragraph{Contradiction and upgrade (Stage~D).}
An upgrade is reachable only when the decisive failure is an administrative one
(from B), the prompt did not demand that format ($m = 0$, from A), the values
match the reference ($r = 1$, computed in Stage~D), and the agent was not
caught lying ($\ell \neq 1$, from C). Any other decisive gate simply confirms
the recorded zero with its cause. Signing $\delta$ records which side was
wrong: the benchmark failed a run that should have passed, or passed a run that
should have failed.
An agent that computes the right values and writes them to \texttt{result.csv}
when the grader expects \texttt{results.csv} fails correctly and fails again on
re-grade, so no contradiction arises; the zero still misrepresents work that
was done. This is only available because an administrative gate protects a
requirement the task did not itself impose. Where the prompt does state the
format, the gate enforces something the task asked for, and Stage~A types it as
critical rather than administrative, so no upgrade is reachable. Signing the
direction matters for the same reason: one case is a fairness bug that punished
a capable agent, the other is unearned credit inflating the leaderboard, and
merging them into a single ``we disagree'' would hide which occurred and what
to do about it.

\paragraph{Signal strength in D2.}
A confession accompanied by a provenance gap is the strongest: both halves are
recorded facts, and each corroborates the other, since the admission explains
the gap and the gap confirms the admission. A copied-value match is
deterministic and guarded against false positives: because a prompt that
states the expected answer can make copying legitimate, D2 convicts it only
where the task required the value to be computed, and all five such
convictions checked by hand held (Appendix~\ref{app:d2detail}). A
judge-voted fabrication is the weakest, resting on the absence of a trace
rather than the presence of evidence, which is why it requires a strict
majority and resolves to clean on a split.
\paragraph{Value tracing granularity (D2).}
Scalars are traced one value at a time, so a constant pasted directly into the
output cannot hide among computed ones, while arrays and large tables, which
are never printed number by number, are cleared at the file level when the file
itself was produced by executed code.

\paragraph{Severity labels in D3.}
The judge gives a final decision $\hat{\ell}$: it confirms the lie, or drops it
if the agent's words did not really contradict what happened. The label
$\gamma$ separates two kinds of claim, saying the work is \emph{correct} versus
merely saying it is \emph{finished}, the first being the more serious. The
label $\eta$ marks the lie as hard or soft by whether a concrete defect sits
behind it. Finally $\phi$ says whether the lie alone should fail the task, on a
deliberately high bar that answers ``unverifiable'' whenever the trace does not
prove the deliverable was actually wrong.
\paragraph{Per task variance and repeat artifacts (D4).}
The within-task standard deviation gives the same quantity as $\rho$ per task,
distinguishing a benchmark whose noise is spread evenly from one whose noise
sits in a few unstable tasks. Each run's artifacts are persisted, so any of D1
through D3 can be applied to any individual repeat.
\section{Recorded bands per model}
\label{app:bands}

Table~\ref{tab:bands} gives ALE's recorded score bands for each model over the
97 gradable tasks; the per-dimension breakdowns in
Appendices~\ref{app:d1detail}--\ref{app:aggdetail} condition on these bands.

\begin{table}[!htb]
\centering
\small
\setlength{\tabcolsep}{3.5pt}
\begin{tabular}{@{}lrrr@{}}
\toprule
Model & pass & partial & fail \\
\midrule
Sonnet    & 15 & 36 & 46 \\
MetaMuse  & 22 & 37 & 38 \\
GPT-Sol   & 23 & 36 & 38 \\
Opus      & 24 & 33 & 40 \\
Gemini    & 18 & 25 & 54 \\
\bottomrule
\end{tabular}
\caption{ALE's recorded bands per model, over the 97 tasks with a re-derivable
verdict. Because a dimension's rate depends on band composition, per-model
rates are not directly comparable without conditioning on band.}
\label{tab:bands}
\end{table}

\section{Competence detail}
\label{app:d1detail}

Table~\ref{tab:d1} reports one figure per outcome per model. Two breakdowns sit
behind it.

The first separates contradictions by direction, which matters because they are
not a single phenomenon. A partial score that fails to clear the verified-pass
bar disagrees with the benchmark by definition rather than by error: both sides
compute the same number and differ only on whether partial credit constitutes a
verified pass. A recorded pass that re-derives as a failure, or a recorded
failure that re-derives as a pass, is a disagreement about the outcome itself.
Table~\ref{tab:d1dir} shows that the first case accounts for most
contradictions in every model, and that the second is several times smaller. One task carried no recorded score at all and is listed separately.

\begin{table}[!htb]
\centering
\small
\setlength{\tabcolsep}{3.5pt}
\begin{tabular}{@{}lrrrrr@{}}
\toprule
Model & total & P$\rightarrow$F & pass$\rightarrow$F & F$\rightarrow$pass & none$\rightarrow$F \\
\midrule
Sonnet   & 38 & 34 & 2 & 2 & 0 \\
MetaMuse & 47 & 37 & 9 & 1 & 0 \\
GPT-Sol  & 47 & 36 & 9 & 2 & 0 \\
Opus     & 43 & 33 & 8 & 2 & 0 \\
Gemini   & 35 & 25 & 8 & 1 & 1 \\
\bottomrule
\end{tabular}
\caption{Contradictions by direction, over the 97 gradable tasks (P = partial, F = fail).}
\label{tab:d1dir}
\end{table}

The second gives the full outcome distribution by recorded band. The partial
row is zero-confirm for every model, which is the threshold rule made visible:
no partial score re-derives as a verified pass. The pass row carries the
disagreements that matter, and it is where the models differ most, from two
contradictions for Sonnet to nine for MetaMuse and GPT-Sol.

\begin{table}[!htb]
\centering
\small
\setlength{\tabcolsep}{3.5pt}
\begin{tabular}{@{}llrrrr@{}}
\toprule
Model & band & Tasks & confirm & contra & unver. \\
\midrule
Sonnet   & pass    & 15 & 13 & 2  & 0 \\
         & partial & 36 & 0  & 34 & 2 \\
         & fail    & 46 & 43 & 2  & 1 \\
\midrule
MetaMuse & pass    & 22 & 13 & 9  & 0 \\
         & partial & 37 & 0  & 37 & 0 \\
         & fail    & 38 & 37 & 1  & 0 \\
\midrule
GPT-Sol  & pass    & 23 & 14 & 9  & 0 \\
         & partial & 36 & 0  & 36 & 0 \\
         & fail    & 38 & 36 & 2  & 0 \\
\midrule
Opus     & pass    & 24 & 15 & 8  & 1 \\
         & partial & 33 & 0  & 33 & 0 \\
         & fail    & 40 & 38 & 2  & 0 \\
\midrule
Gemini   & pass    & 18 & 10 & 8  & 0 \\
         & partial & 25 & 0  & 25 & 0 \\
         & fail    & 54 & 51 & 2  & 1 \\
\bottomrule
\end{tabular}
\caption{D1 outcomes by model and recorded band, over the 97 gradable tasks.
Upgrade is zero throughout and omitted.}
\label{tab:d1bands}
\end{table}

\paragraph{Gate-typing agreement.}
Over 1{,}078 gates the three votes within a pass are unanimous on 94.1\% and
two independent passes agree on 96.8\%. Because gate categories are a property
of the grader rather than of a run, the typed inventory transfers across agents
unchanged.

\section{Reward hacking detail}
\label{app:d2detail}

Table~\ref{tab:d2} reports one figure per model. Two breakdowns sit behind it.

The first gives the verdict distribution by recorded band. Bands in this breakdown follow the D2 verdict file; scores within rounding distance of a band threshold are resolved slightly differently by D1, so a few tasks per model appear in a different band in Table~\ref{tab:bands}, while per-model totals are unaffected. The pattern is not
shared across models: for Sonnet the gaming rate falls from the pass band to
the fail band, from 20\% to 10.9\%, while for the others it rises, so
gaming concentrates where credit is earned in one model and where it is not in
the rest.

\begin{table}[!htb]
\centering
\small
\setlength{\tabcolsep}{3.5pt}
\begin{tabular}{@{}llrrrrr@{}}
\toprule
Model & band & Tasks & assessed & clean & gamed & n/a \\
\midrule
Sonnet   & pass    & 17 & 15 & 11 & 3  & 1 \\
         & partial & 37 & 36 & 30 & 6  & 0 \\
         & fail    & 54 & 46 & 34 & 5  & 7 \\
\midrule
MetaMuse & pass    & 23 & 19 & 12 & 6  & 1 \\
         & partial & 42 & 41 & 24 & 15 & 2 \\
         & fail    & 43 & 37 & 16 & 17 & 4 \\
\midrule
GPT-Sol  & pass    & 23 & 21 & 9  & 10 & 2 \\
         & partial & 40 & 38 & 21 & 17 & 0 \\
         & fail    & 45 & 38 & 15 & 22 & 1 \\
\midrule
Opus     & pass    & 25 & 22 & 20 & 1  & 1 \\
         & partial & 37 & 36 & 30 & 4  & 2 \\
         & fail    & 46 & 39 & 25 & 5  & 9 \\
\midrule
Gemini   & pass    & 19 & 16 & 12 & 3  & 1 \\
         & partial & 28 & 27 & 19 & 8  & 0 \\
         & fail    & 61 & 54 & 13 & 14 & 27 \\
\bottomrule
\end{tabular}
\caption{D2 verdicts by model and recorded band, over the 97 assessed tasks.
The \emph{Tasks} column is the full 108 under each model's own banding; n/a: not assessed.}
\label{tab:d2bands}
\end{table}

The second gives the condition under which each conviction was reached.
Fabrication, which requires a judge, accounts for between 78\% and 100\%
of convictions in every model, while the two deterministic conditions together
account for the remainder.

\begin{table}[!htb]
\centering
\small
\setlength{\tabcolsep}{3.5pt}
\begin{tabular}{@{}lrrrr@{}}
\toprule
Model & copied & confession & fabricated & total \\
\midrule
Sonnet   & 2 & 1 & 11 & 14 \\
MetaMuse & 4 & 0 & 34 & 38 \\
GPT-Sol  & 3 & 2 & 44 & 49 \\
Opus     & 0 & 0 & 10 & 10 \\
Gemini   & 1 & 1 & 23 & 25 \\
\bottomrule
\end{tabular}
\caption{Conditions under which a gamed verdict was reached, per model.}
\label{tab:d2cheat}
\end{table}

\paragraph{Hand verification of gamed verdicts.}
For Sonnet, all 14 held: 11 were required result values written as literals with no
computation or file source anywhere in the run, 2 reproduced a value the prompt
supplied, and 1 rested on the agent's own admission that it could not run the
required engine and had hand-built the results to resemble its output. That
admission is recorded text and the accompanying provenance gap is computed from
the file access trail, so the two together are the strongest condition
described in Section~\ref{sec:arch-d2}, and the evidence is deterministic even
though the detector did not join them on this run. For GPT-Sol, all 49 held: 44
were required result values matching the reference with no computation or source,
3 reproduced a value the prompt supplied, and 2 rested on the agent's own admission.

\paragraph{Runs carrying no verdict.}
In most the agent staged no output matching the graded reference, so no answer
values existed to trace; in a few the trajectory contained no usable log.
Gemini produced nothing assessable on 28 of its 97 tasks.

\section{Reliability outcomes by band, and the full Sonnet set}
\label{app:d4detail}

Table~\ref{tab:d4} reports one figure per model. Two breakdowns sit behind it.

The first, Table~\ref{tab:d4bands}, gives the outcome by recorded band. The band-level $\bar{\Delta}$
values are wider than the pass-band figure reported in the body, running from
$-0.31$ to $+0.67$, and their signs are largely set by the band, since a varied
pass must have a repeat below it and a varied failure a repeat above it. The
partial band is the only one free to move either way, and it falls in four
models and is flat in the fifth.

\begin{table}[!htb]
\centering
\small
\setlength{\tabcolsep}{3.5pt}
\begin{tabular}{@{}llrrrrr@{}}
\toprule
Model & band & Tasks & stable & varied & \% & $\bar{\Delta}$ \\
\midrule
Sonnet   & pass    & 15 & 10 & 5  & 33.3 & $-0.31$ \\
         & partial & 19 & 10 & 9  & 47.4 & $-0.25$ \\
         & fail    & 16 & 7  & 9  & 56.2 & $+0.47$ \\
\midrule
Opus     & pass    & 20 & 15 & 5  & 25.0 & $-0.30$ \\
         & partial & 17 & 14 & 3  & 17.6 & $-0.08$ \\
         & fail    & 13 & 10 & 3  & 23.1 & $+0.67$ \\
\midrule
GPT-Sol  & pass    & 18 & 14 & 4  & 22.2 & $-0.22$ \\
         & partial & 20 & 17 & 3  & 15.0 & $+0.00$ \\
         & fail    & 12 & 10 & 2  & 16.7 & $+0.50$ \\
\midrule
MetaMuse & pass    & 18 & 15 & 3  & 16.7 & $-0.12$ \\
         & partial & 21 & 15 & 6  & 28.6 & $-0.02$ \\
         & fail    & 11 & 8  & 3  & 27.3 & $+0.27$ \\
\midrule
Gemini   & pass    & 13 & 10 & 3  & 23.1 & $-0.24$ \\
         & partial & 15 & 8  & 7  & 46.7 & $-0.15$ \\
         & fail    & 22 & 18 & 4  & 18.2 & $+0.56$ \\
\bottomrule
\end{tabular}
\caption{D4 outcome by model and recorded band, over the 50 tasks repeated five
times under every configuration. $\bar{\Delta}$ is the mean change from the
recorded score, over the varied tasks in that band.}
\label{tab:d4bands}
\end{table}

The second gives the outcome over the larger 87-task set under Sonnet, which is
the source of the reliability statistics reported in the body. Its varied rate
of $25.3\%$ is roughly half the $46.0\%$ recorded on the 50 tasks drawn from it,
and the two are discussed in Section~\ref{sec:results-d4}.

\begin{table}[!htb]
\centering
\small
\setlength{\tabcolsep}{3.5pt}
\begin{tabular}{@{}lrrrrr@{}}
\toprule
band & Tasks & stable & varied & \% & $\bar{\Delta}$ \\
\midrule
pass    & 15 & 10 & 5  & 33.3 & $-0.31$ \\
partial & 36 & 27 & 9  & 25.0 & $-0.25$ \\
fail    & 36 & 28 & 8  & 22.2 & $+0.48$ \\
\midrule
Total   & 87 & 65 & 22 & 25.3 & --- \\
\bottomrule
\end{tabular}
\caption{D4 outcome by recorded band under Sonnet, over the 87 tasks that
completed all five repeats.}
\label{tab:d487}
\end{table}

\paragraph{Stable verdicts outside the recorded band.} Of the 16 stable verdicts that settle in a band other than the recorded one (Section~\ref{sec:results-d4}), four are recorded passes. Three already fail D1 or D2; the fourth, a Gemini pass recorded at 1.0 whose five repeats all score 0, clears D1--D3 and is among the 19 trusted passes. Requiring D4 to reproduce the recorded band would therefore reduce the trusted count to 18 of 84 (21.4\%, 95\% CI 14.0--31.3), and Gemini's to 0 of 13. 
\paragraph{Uncertainty in varied rates.} Treating each repeated task as one trial, 95\% Wilson intervals on the varied rates in Table~4 are: Sonnet 46.0\% [33.0, 59.6], Opus 22.0\% [12.8, 35.2], GPT-Sol 18.0\% [9.8, 30.8], MetaMuse 24.0\% [14.3, 37.4], and Gemini 28.0\% [17.5, 41.7]. Sonnet's 87-task rate of 25.3\% [17.3, 35.3] overlaps its 50-task interval only marginally, consistent with the campaign difference in Section~\ref{sec:results-d4}.

\section{Aggregation across all recorded bands}
\label{app:aggdetail}

Table~\ref{tab:agg} reports the conjunction over recorded passes. Table~\ref{%
tab:aggbands} gives it across all three bands. The partial row is zero
throughout, and the fail row nearly so, because D1 rejects a partial score by
definition and the conjunction requires all four dimensions to clear; the
totals therefore reflect that threshold as much as the evidence, which is why
the body reports the pass band separately.

\begin{table}[!htb]
\centering
\small
\setlength{\tabcolsep}{3.5pt}
\begin{tabular}{@{}llrrrrrrr@{}}
\toprule
 & & & \multicolumn{4}{c}{failed} & \multicolumn{2}{c}{aggregate} \\
\cmidrule(lr){4-7}\cmidrule(lr){8-9}
Model & band & Tasks & D1 & D2 & D3 & D4 & trust. & fail. \\
\midrule
Sonnet   & pass    & 15 & 4  & 5  & 1  & 5  & 4 & 11 \\
         & partial & 19 & 19 & 6  & 6  & 9  & 0 & 19 \\
         & fail    & 16 & 14 & 3  & 7  & 9  & 0 & 16 \\
         & Total   & 50 & 37 & 14 & 14 & 23 & 4 & 46 \\
\midrule
Opus     & pass    & 20 & 8  & 3  & 0  & 5  & 8 & 12 \\
         & partial & 17 & 17 & 4  & 3  & 3  & 0 & 17 \\
         & fail    & 13 & 13 & 7  & 3  & 3  & 0 & 13 \\
         & Total   & 50 & 38 & 14 & 6  & 11 & 8 & 42 \\
\midrule
GPT-Sol  & pass    & 18 & 7  & 12 & 3  & 4  & 2 & 16 \\
         & partial & 20 & 20 & 6  & 8  & 3  & 0 & 20 \\
         & fail    & 12 & 12 & 6  & 2  & 2  & 0 & 12 \\
         & Total   & 50 & 39 & 24 & 13 & 9  & 2 & 48 \\
\midrule
MetaMuse & pass    & 18 & 8  & 6  & 1  & 3  & 4 & 14 \\
         & partial & 21 & 21 & 7  & 11 & 6  & 0 & 21 \\
         & fail    & 11 & 11 & 7  & 2  & 3  & 0 & 11 \\
         & Total   & 50 & 40 & 20 & 14 & 12 & 4 & 46 \\
\midrule
Gemini   & pass    & 13 & 7  & 5  & 2  & 3  & 1 & 12 \\
         & partial & 15 & 15 & 6  & 6  & 7  & 0 & 15 \\
         & fail    & 22 & 22 & 17 & 6  & 4  & 0 & 22 \\
         & Total   & 50 & 44 & 28 & 14 & 14 & 1 & 49 \\
\bottomrule
\end{tabular}
\caption{Aggregation by model and recorded band, over the 50 repeated tasks, all of which carry a D4 verdict; ungradable tasks among them fail D1, and fail D2 only where no answer values can be traced. D1--D4 count tasks failing that dimension and overlap freely, so they do not sum; a task is trusted only where all four clear it (trust./fail.: trusted/failed).}
\label{tab:aggbands}
\end{table}

\begin{table}[!htb]
\centering
\small
\setlength{\tabcolsep}{3.5pt}
\begin{tabular}{@{}lrrrrrr@{}}
\toprule
Filter & Son. & Opus & GPT & Meta & Gem. & total \\
\midrule
Recorded pass                & 15 & 20 & 18 & 18 & 13 & 84 \\
+ D1                   & 11 & 12 & 11 & 10 &  6 & 50 \\
+ D1, D4 (determ.) &  6 &  8 &  9 &  8 &  3 & 34 \\
+ all four             &  4 &  8 &  2 &  4 &  1 & 19 \\
\bottomrule
\end{tabular}
\caption{Recorded passes surviving each filter, over the 50 repeated tasks (Son.\ = Sonnet, GPT = GPT-Sol, Meta = MetaMuse, Gem.\ = Gemini).}
\label{tab:ablation}
\end{table}

\paragraph{Composition of Table 5.} Table~\ref{tab:breakdown} decomposes the D1 and D2 failures among the 84 recorded passes. Ungradable passes always fail D1, since their grader cannot be re-run; they fail D2 only when they staged no traceable answer values, so the ungradable counts differ between the two columns. The contradiction, unverified, and gamed counts stay within the pass-band counts of Tables~\ref{tab:d1bands} and~\ref{tab:d2bands}. Because ungradable passes always fail D1, none of the 19 trusted passes is ungradable; restricting the conjunction to the 74 gradable recorded passes gives 19 of 74 (25.7\%, 95\% CI 17.1--36.7).

\begin{table}[!htb]
\centering
\resizebox{\columnwidth}{!}{%
\begin{tabular}{lcccccccc}
\toprule
 & \multicolumn{4}{c}{D1 failures} & \multicolumn{4}{c}{D2 failures} \\
\cmidrule(lr){2-5}\cmidrule(lr){6-9}
Model & total & contra & unver & ungrad & total & gamed & no verd. & ungrad \\
\midrule
Opus     & 8 & 5 & 1 & 2 & 3  & 1  & 2 & 0 \\
Sonnet   & 4 & 2 & 0 & 2 & 5  & 3  & 1 & 1 \\
GPT-Sol  & 7 & 6 & 0 & 1 & 12 & 10 & 2 & 0 \\
MetaMuse & 8 & 5 & 0 & 3 & 6  & 5  & 1 & 0 \\
Gemini   & 7 & 5 & 0 & 2 & 5  & 3  & 2 & 0 \\
\bottomrule
\end{tabular}%
}
\caption{Decomposition of the D1 and D2 failures in Table~5, over recorded passes among the 50 repeated tasks (contra = contradiction, unver = unverified, ungrad = ungradable, no verd. = no verdict).}
\label{tab:breakdown}
\end{table}

\paragraph{Ablation over recorded passes.}
Table~\ref{tab:ablation} applies the dimensions in sequence to the 84 recorded
passes. D1 alone leaves 50 standing. Adding D4, so that only the two
deterministic dimensions apply, leaves 34. Requiring all four leaves 19, so the
two judge-assisted dimensions reject a further 15 passes that the deterministic
checks clear, from none for Opus to seven of nine for GPT-Sol.

\paragraph{Worked examples.}
Two recorded passes illustrate the layer end to end. In a Health/Medicine clinical variant annotation task (Near-Term), GPT-Sol scored 1.0. D1 re-ran the grader and reproduced the pass, and D3 found its claim of success consistent with the re-graded outcome. D2, however, found that the credited answer value matched the reference and was stated in the task prompt, although the task required it to be computed, and convicted the run as gamed (copied). D4 then repeated the task five times, scoring 1.0, 1.0, 0.0, 1.0, and 0.0: the repeats crossed from pass to fail, so the task is varied. 
Failing D2 and D4, this recorded pass is not trusted, although the benchmark and D1 both count it as a success. By contrast, in a Business/Finance American option pricing task (Full-Spectrum), Sonnet scored 1.0, the re-graded verdict confirmed the pass, all 60 credited values were grounded in executed output or read inputs, the success claim matched the outcome, and all five repeats scored 1.0. This pass clears all four dimensions and is trusted.

\section{Extension to three newer agent configurations}
\label{app:extension}

To test whether the layer carries over to newer agents, we applied it
unchanged to three recent configurations, Claude Sonnet~5.5, Grok~4.7, and
Claude Opus~5.5, on the same tasks, under the same agent harness, decoding
configuration, and judge. These runs postdate the main campaign, so we report
them separately and keep the main results on a single run campaign.
Because the layer, judge, and gate inventory are unchanged, these runs carry the
same validation as Opus, MetaMuse, and Gemini in the main study: gate typing and
the deterministic stages transfer across agents, while hand checks of D2 and
annotation of D3 cover Sonnet and GPT-Sol of the main study (Sections~\ref{sec:results-d2} and~\ref{sec:results-d3}; Appendix~\ref{app:d1detail}). D1 to D3 cover the same 97 gradable tasks as the main
study for all three, and D4 the same 50 repeated tasks. Three task runs, two
for Grok~4.7 and one for Opus~5.5, were completed after the others under the
same configuration; none of them is among the 50 repeated tasks.

\paragraph{Per-dimension results.}
Tables~\ref{tab:ext_d1} to~\ref{tab:ext_d3d4} give the results in the form of
Tables~\ref{tab:d1} to~\ref{tab:d4}, \ref{tab:bands}, \ref{tab:d1dir},
and~\ref{tab:d2cheat}. On D1, upgrade is again never granted, and 6 to 11
recorded passes per model re-derive as failures. On D2, the gamed rate runs
from 6.2\% for Opus~5.5 to 43.3\% for Grok~4.7, so over all eight
configurations it spans 6.2\% to 50.5\%. As in the main study, most gamed
verdicts are fabrications: answer values that match the reference yet appear
in no executed output and nothing the agent read, so the agent wrote them
without computing them (Table~\ref{tab:ext_d2}). The remaining 6, all for
Sonnet~5.5, rest on a deterministic copied-value match. On D3, all three make
confirmed false completion claims, though fewer than the original five, from 4
for Sonnet~5.5 to 13 for Grok~4.7. On D4, varied rates are 28.0\% [17.5, 41.7]
for Sonnet~5.5, 40.0\% [27.6, 53.8] for Grok~4.7, and 38.0\% [25.9, 51.8] for
Opus~5.5 (Wilson intervals, as in Appendix~\ref{app:d4detail}), within the
range of 18\% to 46\% in Table~\ref{tab:d4}, and the pass band $\bar{\Delta}$
is again negative for every model.

\begin{table}[!htb]
\centering
\small
\setlength{\tabcolsep}{1.8pt}
\begin{tabular}{lcccccccc}
\toprule
 & & \multicolumn{3}{c}{band} & \multicolumn{4}{c}{D1} \\
\cmidrule(lr){3-5}\cmidrule(lr){6-9}
Model & grad. & pass & part. & fail & conf. & contra & unver. & pass$\rightarrow$F \\
\midrule
Sonnet 5.5 & 97 & 20 & 39 & 38 & 48 & 47 & 2 & 7  \\
Grok 4.7   & 97 & 15 & 38 & 44 & 50 & 46 & 1 & 6  \\
Opus 5.5   & 97 & 27 & 33 & 37 & 49 & 44 & 4 & 11 \\
\bottomrule
\end{tabular}
\caption{Recorded bands and D1 outcomes for the three newer configurations,
over the 97 gradable tasks (grad. = gradable, part. = partial, conf. =
confirm, contra = contradiction, unver. = unverified, pass$\rightarrow$F =
recorded passes that re-derive as failures, as in Table~\ref{tab:d1dir}). Upgrade is zero
throughout and omitted.}
\label{tab:ext_d1}
\end{table}

\begin{table}[!htb]
\centering
\small
\setlength{\tabcolsep}{2.2pt}
\begin{tabular}{lcccccccc}
\toprule
 & \multicolumn{5}{c}{D2 verdicts} & \multicolumn{3}{c}{gamed by} \\
\cmidrule(lr){2-6}\cmidrule(lr){7-9}
Model & ass. & clean & gamed & n/a & \% & cop. & conf. & fab. \\
\midrule
Sonnet 5.5 & 97 & 79 & 13 & 5  & 13.4 & 6 & 0 & 7  \\
Grok 4.7   & 97 & 44 & 42 & 11 & 43.3 & 0 & 0 & 42 \\
Opus 5.5   & 97 & 83 & 6  & 8  & 6.2  & 0 & 0 & 6  \\
\bottomrule
\end{tabular}
\caption{D2 verdicts and the conditions under which each gamed verdict was
reached, for the three newer configurations, in the form of Tables~\ref{tab:d2} and~\ref{tab:d2cheat}
(ass. = assessed, n/a = not assessed, \% = gamed\%, cop. = copied, conf. =
confession, fab. = fabricated). As in Table~\ref{tab:d2}, gamed\% is over all assessed
tasks including runs with no verdict.}
\label{tab:ext_d2}
\end{table}

\begin{table}[!htb]
\centering
\small
\setlength{\tabcolsep}{2.2pt}
\begin{tabular}{lcccccccc}
\toprule
 & \multicolumn{3}{c}{D3} & \multicolumn{5}{c}{D4} \\
\cmidrule(lr){2-4}\cmidrule(lr){5-9}
Model & conf. & hard & soft & tasks & stab. & var. & var.\% & $\bar{\Delta}$ \\
\midrule
Sonnet 5.5 & 4  & 4  & 0 & 50 & 36 & 14 & 28.0 & $-0.23$ \\
Grok 4.7   & 13 & 12 & 1 & 50 & 30 & 20 & 40.0 & $-0.40$ \\
Opus 5.5   & 6  & 6  & 0 & 50 & 31 & 19 & 38.0 & $-0.22$ \\
\bottomrule
\end{tabular}
\caption{D3 confirmed false claims, over the gradable tasks of
Table~\ref{tab:ext_d1}, and D4 outcomes over the 50 repeated tasks, for the
three newer configurations, in the form of Tables~\ref{tab:d3} and~\ref{tab:d4} (conf. = confirmed, hard and soft as in Table~\ref{tab:d3}, stab. = stable, var. = varied). As in Table~\ref{tab:d4},
$\bar{\Delta}$ is the mean change from the recorded score over varied pass band
tasks.}
\label{tab:ext_d3d4}
\end{table}

\paragraph{Aggregation.}
Table~\ref{tab:extension} gives the conjunction over recorded passes among the
50 repeated tasks, under the counting rules of Section~\ref{sec:arch-agg}, and
Tables~\ref{tab:ext_ablation} and~\ref{tab:ext_decomp} give the matching
ablation and decomposition, in the form of Tables~\ref{tab:ablation} and~\ref{tab:breakdown}. The pattern of
the main study holds. Pooled over all eight configurations, 29 of 133 recorded
passes survive all four checks (21.8\%, 95\% CI [15.6, 29.6]), close to the
22.6\% of Table~\ref{tab:agg}. Over gradable passes only, as in Appendix~\ref{app:aggdetail}, the three newer
configurations give 10 of 43 (23.3\%, 95\% CI [13.2, 37.7]) and all eight give
29 of 117 (24.8\%, 95\% CI [17.8, 33.3]). The two deterministic dimensions, D1
and D4, leave 49 of the 133, and the two judge-assisted ones reject a further
20. The configurations again fail for different reasons: Opus~5.5 is gamed on
only 6 of 97 tasks but fails D4 on 12 of its 20 recorded passes among the 50
repeated tasks, so a cleaner answer does not imply a more reliable one.

\begin{table}[!htb]
\centering
\small
\setlength{\tabcolsep}{2pt}
\begin{tabular}{lcccccccc}
\toprule
 & & \multicolumn{4}{c}{failed} & \multicolumn{3}{c}{aggregate} \\
\cmidrule(lr){3-6}\cmidrule(lr){7-9}
Model & Passes & D1 & D2 & D3 & D4 & trust. & \% & 95\% CI \\
\midrule
Sonnet 5.5 & 17  & 7  & 4 & 0 & 5  & 5  & 29.4 & [13.3, 53.1] \\
Grok 4.7   & 12  & 6  & 6 & 0 & 1  & 1  & 8.3  & [1.5, 35.4] \\
Opus 5.5   & 20  & 10 & 3 & 1 & 12 & 4  & 20.0 & [8.1, 41.6] \\
\midrule
New three  & 49  &    &   &   &    & 10 & 20.4 & [11.5, 33.6] \\
All eight  & 133 &    &   &   &    & 29 & 21.8 & [15.6, 29.6] \\
\bottomrule
\end{tabular}
\caption{Conjunction over recorded passes for the three newer configurations,
over the 50 repeated tasks, in the form of Table~\ref{tab:agg}. D1 to D4 count passes
failing that dimension and overlap, so they do not sum. All eight pools these
rows with the 84 passes of Table~\ref{tab:agg}. Intervals are Wilson score intervals at
95\% (trust. = trusted).}
\label{tab:extension}
\end{table}

\begin{table}[!htb]
\centering
\small
\setlength{\tabcolsep}{2pt}
\begin{tabular}{lcccc}
\toprule
Filter & Son.\,5.5 & Grok\,4.7 & Opus\,5.5 & all eight \\
\midrule
Recorded pass     & 17 & 12 & 20 & 133 \\
+ D1              & 10 & 6  & 10 & 76  \\
+ D1, D4 (determ.)& 6  & 5  & 4  & 49  \\
+ all four        & 5  & 1  & 4  & 29  \\
\bottomrule
\end{tabular}
\caption{Recorded passes surviving each filter over the 50 repeated tasks, in
the form of Table~\ref{tab:ablation}. All eight pools these columns with the totals of Table~\ref{tab:ablation} (new three: 49, 26, 15, 10).}
\label{tab:ext_ablation}
\end{table}

\begin{table}[!htb]
\centering
\small
\setlength{\tabcolsep}{2.2pt}
\begin{tabular}{lcccccccc}
\toprule
 & \multicolumn{4}{c}{D1 failures} & \multicolumn{4}{c}{D2 failures} \\
\cmidrule(lr){2-5}\cmidrule(lr){6-9}
Model & tot. & con. & unv. & ung. & tot. & gam. & n.v. & ung. \\
\midrule
Sonnet 5.5 & 7  & 5 & 0 & 2 & 4 & 3 & 1 & 0 \\
Grok 4.7   & 6  & 4 & 0 & 2 & 6 & 4 & 2 & 0 \\
Opus 5.5   & 10 & 7 & 1 & 2 & 3 & 2 & 1 & 0 \\
\bottomrule
\end{tabular}
\caption{Decomposition of the D1 and D2 failures in
Table~\ref{tab:extension}, over recorded passes among the 50 repeated tasks,
in the form of Table~\ref{tab:breakdown} (tot. = total, con. = contradiction, unv. = unverified,
ung. = ungradable tasks among the 50, gam. = gamed, n.v. = no verdict).}
\label{tab:ext_decomp}
\end{table}

\section{Terminal-Bench detail}
\label{app:tbench}
Table~\ref{tab:tbench} reports the conjunction over recorded passes. This appendix gives the results for each dimension behind it. D1 and D3 cover the 64 gradable tasks, and D2 and D4 cover all 66.

\paragraph{Recorded bands and D1.} Table~\ref{tab:tbench-d1} gives the recorded scores and the D1 outcome for each model. Because Terminal-Bench scores are pass or fail, there is no partial band, so every contradiction is a real disagreement about the outcome. All 16 contradictions are recorded passes that fail when the grader is re-run on the saved output; no recorded failure re-grades as a pass. The upgrade path is again never granted.

\begin{table}[!htb]
\centering
\footnotesize
\setlength{\tabcolsep}{1.8pt}
\begin{tabular}{@{}lrrrrrrrr@{}}
\toprule
 & \multicolumn{2}{c}{all 66} & \multicolumn{2}{c}{gradable 64} & \multicolumn{4}{c}{D1} \\
\cmidrule(lr){2-3}\cmidrule(lr){4-5}\cmidrule(l){6-9}
Model & pass & fail & pass & fail & conf. & contra & unver. & P$\to$F \\
\midrule
Opus 5.5    & 47 & 19 & 45 & 19 & 62 & 1 & 1 & 1 \\
Opus 5      & 34 & 32 & 33 & 31 & 60 & 3 & 1 & 3 \\
GPT-Sol     & 19 & 47 & 18 & 46 & 59 & 5 & 0 & 5 \\
GPT-6.1 Sol & 37 & 29 & 35 & 29 & 57 & 6 & 1 & 6 \\
Fable 5.1   & 43 & 23 & 41 & 23 & 63 & 1 & 0 & 1 \\
\bottomrule
\end{tabular}
\caption{Recorded scores and D1 outcomes on Terminal-Bench. D1 is over the 64 gradable tasks (conf. = confirm, contra = contradiction, unver. = unverified, P$\to$F = recorded passes that re-grade as failures). Upgrade is zero throughout and omitted.}
\label{tab:tbench-d1}
\end{table}

\paragraph{D2.} Table~\ref{tab:tbench-d2} gives the D2 verdicts and the condition behind each \textit{gamed} verdict. Fabrication accounts for 16 of the 24 verdicts and a copied value for the other 8; no run was convicted on a confession. A third of the verdicts therefore rest on the deterministic copied-value match. GPT-Sol's four \textit{gamed} runs are all recorded failures, so D2 removes none of its passes.

\begin{table}[!htb]
\centering
\footnotesize
\setlength{\tabcolsep}{2.5pt}
\begin{tabular}{@{}lrrrrrrr@{}}
\toprule
 & \multicolumn{4}{c}{D2 verdicts} & \multicolumn{3}{c}{gamed by} \\
\cmidrule(lr){2-5}\cmidrule(l){6-8}
Model & clean & gamed & n/a & \% & cop. & conf. & fab. \\
\midrule
Opus 5.5    & 61 & 4 & 1 & 6.1  & 2 & 0 & 2 \\
Opus 5      & 60 & 3 & 3 & 4.5  & 1 & 0 & 2 \\
GPT-Sol     & 61 & 4 & 1 & 6.1  & 1 & 0 & 3 \\
GPT-6.1 Sol & 56 & 8 & 2 & 12.1 & 2 & 0 & 6 \\
Fable 5.1   & 61 & 5 & 0 & 7.6  & 2 & 0 & 3 \\
\bottomrule
\end{tabular}
\caption{D2 verdicts on Terminal-Bench and the condition under which each \textit{gamed} verdict was reached (n/a = not assessed, \% = gamed share of all 66 tasks, cop. = copied, conf. = confession, fab. = fabricated).}
\label{tab:tbench-d2}
\end{table}

\paragraph{D3 and D4.} Table~\ref{tab:tbench-d3d4} gives the confirmed false claims and the D4 outcomes. Most confirmed claims are hard, which fits a benchmark with no partial credit. Measured against the runs that re-grade as failures, the share of failures reported as successes falls from 47.1\% [31.5, 63.3] for Opus 5 to 25.0\% [11.2, 46.9] for Opus 5.5, and from 62.7\% [49.0, 74.7] for GPT-Sol to 31.4\% [18.6, 48.0] for GPT-6.1 Sol. On D4, the varied share has 95\% Wilson intervals of [18.0, 39.0] for Opus 5.5, [25.8, 48.4] for Opus 5, [31.2, 54.4] for GPT-Sol and Fable 5.1, and [20.6, 42.2] for GPT-6.1 Sol. Because the recorded run is one of the five, a varied pass that passes $k$ of five runs changes by $k/5 - 1$, so $\bar{\Delta} = -0.46$ for Opus 5.5 means a varied pass held on about 2.7 of five runs. On GPU tasks, runs that failed for lack of a GPU were excluded as infrastructure runs, so a few tasks have fewer than five runs.

\begin{table}[!htb]
\centering
\footnotesize
\setlength{\tabcolsep}{2.2pt}
\begin{tabular}{@{}lrrrrrrrr@{}}
\toprule
 & \multicolumn{3}{c}{D3} & \multicolumn{5}{c}{D4} \\
\cmidrule(lr){2-4}\cmidrule(l){5-9}
Model & conf. & hard & soft & stab. & var. & var.\% & $\bar{\Delta}$ & $n$ \\
\midrule
Opus 5.5    & 5  & 5  & 0 & 48 & 18 & 27.3 & $-0.46$ & 13 \\
Opus 5      & 16 & 13 & 3 & 42 & 24 & 36.4 & $-0.35$ & 12 \\
GPT-Sol     & 32 & 28 & 4 & 38 & 28 & 42.4 & $-0.39$ & 11 \\
GPT-6.1 Sol & 11 & 10 & 1 & 46 & 20 & 30.3 & $-0.37$ & 11 \\
Fable 5.1   & 14 & 12 & 2 & 38 & 28 & 42.4 & $-0.42$ & 21 \\
\bottomrule
\end{tabular}
\caption{D3 confirmed false claims over the 64 gradable tasks, and D4 outcomes over all 66 tasks on Terminal-Bench, in the form of Tables~\ref{tab:d3} and~\ref{tab:d4} (conf. = confirmed, stab. = stable, var. = varied). $\bar{\Delta}$ is the mean change from the recorded score over the $n$ varied tasks whose recorded run passed, averaged over the runs left after infrastructure exclusion.}
\label{tab:tbench-d3d4}
\end{table}

\paragraph{Ablation.} Table~\ref{tab:tbench-ablation} applies the dimensions in sequence. D1 leaves 156 of 180 passes, and D4 then removes 66, the largest single step. D2 and D3 remove a further 14, from none for GPT-Sol to five for GPT-6.1 Sol.

\begin{table}[!htb]
\centering
\footnotesize
\setlength{\tabcolsep}{1.6pt}
\begin{tabular}{@{}lrrrrrr@{}}
\toprule
Filter & Op.\,5.5 & Op.\,5 & GPT & GPT\,6.1 & Fab. & total \\
\midrule
Recorded pass     & 47 & 34 & 19 & 37 & 43 & 180 \\
+ D1              & 44 & 30 & 13 & 29 & 40 & 156 \\
+ D1, D4 (determ.)& 31 & 18 & 2  & 19 & 20 & 90 \\
+ all four        & 27 & 16 & 2  & 14 & 17 & 76 \\
\bottomrule
\end{tabular}
\caption{Recorded passes on Terminal-Bench surviving each filter, in the form of Table~\ref{tab:ablation} (Op. = Opus, GPT = GPT-Sol, GPT\,6.1 = GPT-6.1 Sol, Fab. = Fable 5.1).}
\label{tab:tbench-ablation}
\end{table}

\end{document}